\documentclass[sigconf]{acmart}
\AtBeginDocument{%
  \providecommand\BibTeX{{%
    \normalfont B\kern-0.5em{\scshape i\kern-0.25em b}\kern-0.8em\TeX}}}

\copyrightyear{2023}
\acmYear{2023}
\setcopyright{rightsretained}
\acmConference[MM '23]{Proceedings of the 31st ACM International Conference on Multimedia}{October 29-November 3, 2023}{Ottawa, ON, Canada}
\acmBooktitle{Proceedings of the 31st ACM International Conference on Multimedia (MM '23), October 29-November 3, 2023, Ottawa, ON, Canada}
\acmDOI{10.1145/3581783.3612248}
\acmISBN{979-8-4007-0108-5/23/10}

\makeatletter
\gdef\@copyrightpermission{
  \begin{minipage}{0.3\columnwidth}
   \href{https://creativecommons.org/licenses/by/4.0/}{\includegraphics[width=0.90\textwidth]{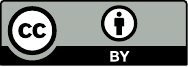}}
  \end{minipage}\hfill
  \begin{minipage}{0.7\columnwidth}
   \href{https://creativecommons.org/licenses/by/4.0/}{This work is licensed under a Creative Commons Attribution International 4.0 License.}
  \end{minipage}
  \vspace{5pt}
}
\makeatother

\RequirePackage{algorithm}
\RequirePackage{algorithmic}
\usepackage{microtype}

\usepackage{epsfig}
\usepackage{graphicx}
\usepackage{bm}
\usepackage{enumitem}
\usepackage{multirow}
\usepackage{amsmath}
\usepackage{csquotes}
\usepackage{amsfonts}
\usepackage{multicol}
\usepackage{multirow}
\usepackage{booktabs}
\def\eqref#1{equation~\ref{#1}}

\def\1{\bm{1}}

\DeclareMathAlphabet{\mathsfit}{\encodingdefault}{\sfdefault}{m}{sl}
\SetMathAlphabet{\mathsfit}{bold}{\encodingdefault}{\sfdefault}{bx}{n}

\newcommand{\Amat}{{\bf A}}

\newcommand{\Cmat}{{\bf C}}

\newcommand{\Qmat}{{\bf Q}}

\newcommand{\Tmat}{{\bf T}}

\newcommand{\av}{{\boldsymbol a}}

\newcommand{\bv}{{\boldsymbol b}}

\newcommand{\knumv}[1]{{\boldsymbol k_{#1}}}

\definecolor{mydarkblue}{rgb}{0,0.08,1}
\definecolor{mydarkgreen}{rgb}{0.02,0.6,0.02}
\definecolor{mydarkred}{rgb}{0.8,0.02,0.02}
\definecolor{mydarkorange}{rgb}{0.40,0.2,0.02}
\definecolor{mypurple}{RGB}{111,0,255}
\definecolor{myred}{rgb}{1.0,0.0,0.0}
\definecolor{mygold}{rgb}{0.75,0.6,0.12}
\definecolor{myblue}{rgb}{0,0.2,0.8}
\definecolor{mydarkgray}{rgb}{0.66,0.66,0.66}

\newcommand\footnoteref[1]{\protected@xdef\@thefnmark{\ref{#1}}\@footnotemark}

\acmSubmissionID{2394}

\ccsdesc[500]{Computing methodologies~Natural language processing}
\ccsdesc[500]{Computing methodologies~Computer vision}
\ccsdesc[500]{Theory of computation~Theory and algorithms for application domains}

\begin{document}

\title[Grounded Language Learning with Multimodal Partial Alignment]{Expand BERT Representation with Visual Information via Grounded Language Learning with Multimodal Partial Alignment}


\author{Cong-Duy Nguyen}
\authornote{\label{note1}Both authors contributed equally to this research.}
\affiliation{
  \institution{Nanyang Technological University}
  \country{Singapore}
}
\email{nguyentr003@ntu.edu.sg}

\author{The-Anh Vu-Le}
\authornotemark[1]
\affiliation{
  \institution{University of Illinois Urbana-Champaign
  \country{USA}}
}
\email{vltanh@illinois.edu}

\author{Thong Nguyen}
\affiliation{%
  \institution{National University of Singapore}
  \country{Singapore}
}
\email{e0998147@u.nus.edu}

\author{Tho Quan}
\affiliation{%
  \institution{Ho Chi Minh City University of Technology (HCMUT), VNU-HCM}
  \country{Vietnam}
}
\email{qttho@hcmut.edu.vn}

\author{Luu Anh Tuan}
\affiliation{
 \institution{Nanyang Technological University}
 \country{Singapore}
}
\email{anhtuan.luu@ntu.edu.sg}


\begin{abstract}
  Language models have been supervised with both language-only objective and visual grounding in existing studies of visual-grounded language learning. However, due to differences in the distribution and scale of visual-grounded datasets and language corpora, the language model tends to mix up the context of the tokens that occurred in the grounded data with those that do not. As a result, during representation learning, there is a mismatch between the visual information and the contextual meaning of the sentence. To overcome this limitation, we propose GroundedBERT - a grounded language learning method that enhances the BERT representation with visually grounded information. GroundedBERT comprises two components: (i) the original BERT which captures the contextual representation of words learned from the language corpora, and (ii) a visual grounding module which captures visual information learned from visual-grounded datasets. Moreover, we employ Optimal Transport (OT), specifically its partial variant, to solve the fractional alignment problem between the two modalities. Our proposed method significantly outperforms the baseline language models on various language tasks of the GLUE and SQuAD datasets.
\end{abstract}



\keywords{Grounded Language Learning, Optimal Transport}



\maketitle

\section{Introduction}
\begin{figure*}[t]
\centering
\includegraphics[width=0.9\textwidth] {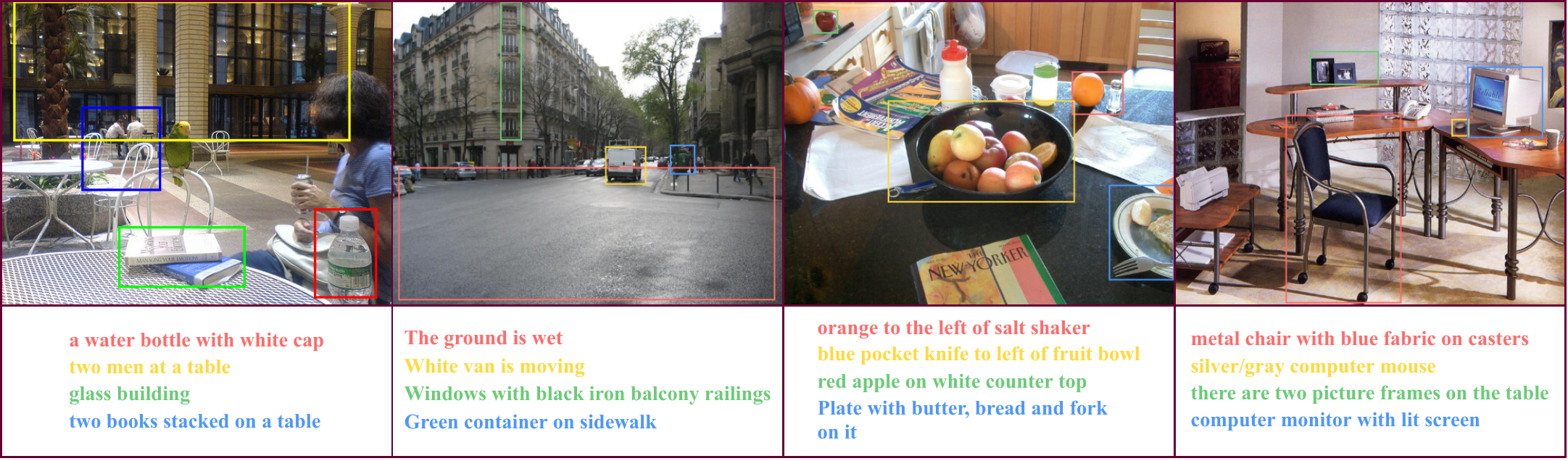}
\caption{ Example image-caption pairs in Visual Genome dataset \cite{krishnavisualgenome}}
\label{fig:VGexample}
\vspace{-5pt}
\end{figure*}

\label{sec:intro}

Grounded language learning is concerned with learning the meaning of language as it applies to the real world. Humans, especially children, learn language from not only pure textual information but also different modalities such as vision and audio, which contain rich information that cannot be captured by text alone \cite{sachs1981language,ogrady2005,vigliocco2014language}. However, many traditional language models are learned only from textual corpora \cite{devlin2019bert, GPT}. They have the limitation in learning complex semantics that require the combination of signals in data through cross-referencing and synthesis.


Recently, there have been many studies trying to improve the language representation with visual information \cite{lazaridou2015combining, collell2017imagined, kiela2018learning, wang2019holistic, wei2022audio, wei2023multi, wei2024learning, bordes2020incorporating, vokenization}. In those attempts, they update the weights of the language encoder using the visual objective together with the pure language-based objective during pretraining. However, there is usually a huge gap in the distribution and quantity of word tokens between visual datasets and language corpora. For example, in Table \ref{table:stat}, the Book Corpus and Wikipedia, two conventional language corpora, contain billions of words with millions of unique tokens, while MS COCO, a common visual-grounded dataset, contains only $6$ million words and $44$ thousand unique tokens. Therefore, during visual-grounded learning, only the tokens from the visual datasets are updated while the majority of the tokens are not equipped with visual information. However, during pretraining, those tokens with and without information from the images will be mixed up in the same context of the sentence, confusing the contextual learning process.

\begin{table}[t]
\caption{Statistics of some common datasets used in visual grounded language learning task.}
\centering
\resizebox{0.44\textwidth}{!}{
\begin{tabular}{@{}l|cc|c@{}}
\toprule
 & \textbf{Book Corpus} & \textbf{Wikipedia} & \textbf{MS COCO}\\
\midrule
\# of words & 985M & 2471M & 6M \\
\# of sentences & 74M & 113M & 616K \\
\# of unique words & 1M & 8M & 44K \\
\bottomrule
\end{tabular} }

\label{table:stat}
\end{table}



Moreover, previous attempts compressed the entire image into one vector as a global representation and then matched it to the paired caption. However, as shown by the samples picked from the Visual Genome \cite{krishnavisualgenome} dataset in Figure \ref{fig:VGexample}, many of the captions only describe local regions in the corresponding image. Thus, using a global representation vector can distract the encoder from capturing local information, making it difficult for the model to align between modalities. As a solution to this issue, we use the Vision Transformer~\cite{ViT} model as the visual encoder to store local information in patch embeddings.

Additionally, aligning information from different modalities is a crucial phase in vision-language representation learning because it is how two sources of information are combined. There are existing researches that use Optimal Transport to solve this alignment problem. Uniter \cite{chen2019uniter} and ViLT \cite{vilt} used the OT-based distance as a pretraining objective, while Graph Optimal Transport \cite{GraphOT} considered two OT distance: Wasserstein distance (WD) and Gromov-Wasserstein distance (GWD) for cross-domain alignment in Visual Question Answering. Nevertheless, the classical optimal transportation problem seeks a transportation map that satisfies marginal constraints, requiring masses from all sources to be moved to all destinations. In some cases, we want only a fraction of masses to be carried, making this requirement restrictive. For instance, as stated above, the caption only describes a part of the image. To get a more flexible alignment, we propose to adapt the Partial Optimal Transport variant to align between the modalities.


Our contributions can be summarized as:
\begin{itemize}
  \item We propose GroundedBERT - a grounded language representation that extends the BERT representation with visual information. The visual-grounded representation is first learned from the text-image pairs and then concatenated with the original BERT representation to form a unified visual-textual representation.
  \item We use patch embeddings from Vision Transformer to maintain local information of the image instead of a single global representation. We also adapt Partial Optimal Transport to align between the two modalities.
  \item We conduct extensive experiments on various language downstream tasks on the GLUE and SQuAD datasets. Empirical result shows that we significantly outperforms the baselines on these tasks.
\end{itemize}

\begin{figure*}[t]
\centering
\includegraphics[width=0.9\textwidth] {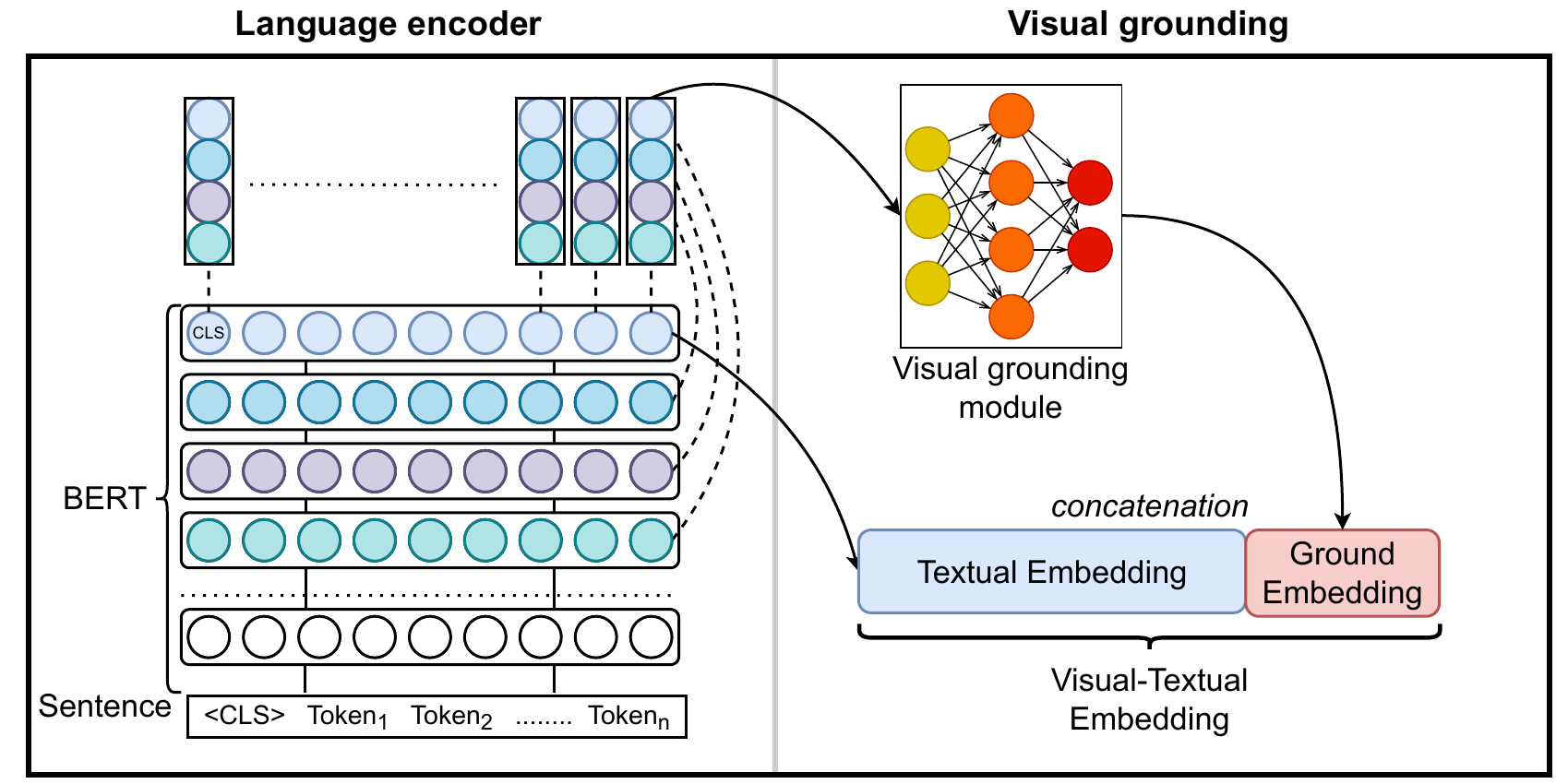}
\caption{Implementation of our GroundedBERT. The model consists of two components, i.e. Language encoder and Visual grounding part. The new representation of language model combines of Textual embedding and Visual embedding.}
\label{fig:vslm}
\vspace{-5pt}
\end{figure*}

\section{Related Work}
\label{sec:related}

Over the past decades, many approaches have been proposed to learn language representation. Skip-gram \cite{mikolov2013distributed}, GLOVE \cite{pennington2014glove} were proposed to learn word representations. On the other hand, FastSent \cite{fastsent}, QuickThought \cite{quickthought}, SkipThought \cite{skip_thought}, Sentence-BERT \cite{SentBert}, or \cite{le2014distributed, conneau2017supervised} tried to learn the sentence representations. Recently, many language models such as  ELMo~\cite{peters2018deep}, BERT~\cite{devlin2019bert}, RoBERTa~\cite{liu2019roberta}, XLNet~\cite{yang2019xlnet}, GPT~\cite{GPT}, ELECTRA~\cite{clark2019electra}, ALBERT~\cite{lan2019albert} were proposed to learn the contextual representation. However, these studies learn the language representation on only textual corpora.

In recent years, many vision-and-language pretrained models have been proposed to build joint cross-modal representations and focus on vision-and-language tasks such as visual question answering and natural language for visual reasoning \cite{li2019visualbert, chen2019uniter, su2019vl}. While \cite{zhou2019unified, li2020oscar} used only one cross-modal Transformer for learning, \cite{lu2019vilbert, tan2019lxmert} proposed to use two single-modal Transformers and one cross-modal Transformer. Pretraining tasks such as masked language model and masked visual-feature classification were used in those studies to learn the vision-and-language representation.

Advanced machine learning algorithms such as the Contrastive Learning framework have been applied to the natural language processing and computer vision~\cite{thong2021contrastive, pan2021i, pan2021c, ContrastiveVLP, nguyen2022adaptive, nguyen2023gradient}. Optimal Transport has also been extensively used in many natural language processing tasks and also the integration of vision and language fields, for example, Cross-Lingual Abstractive Summarization \cite{thong2021cross}, machine translation \cite{chen2019}, Vision and language pretraining \cite{vilt, chen2019uniter}, Visual Question Answering \cite{GraphOT}, etc. Nevertheless, the application of the variants of OT has been less attractive in vision-and-language research.

There are many works on grounded language learning \cite{expgroundslang, Unsupervisedalignment, Visuallygrounded} having been introduced in the past few years. On the other hand, there are few attempts to improve language representation with visual information. \citep{lazaridou2015combining} introduced multimodal skip-gram models (MMSKIP-GRAM) taking visual information into account. \cite{collell2017imagined} proposed IMAGINET which consists of GRU networks and tried to predict the visual representation and the next word in the sentence. \cite{kiela2018learning} was similar to IMAGINET but they used a bi-directional LSTM for sentence encoder. Moreover, it aimed to predict both the visual feature and the other captions given one caption. \cite{bordes2020incorporating} proposed an intermediate space called the grounded space and learns the visual and textual representation with cluster information and perceptual information. \cite{vokenization} introduced the concept of vokenization and pretrained the language model with an additional voken-classification task.

\section{Methodology}
\label{sec:method}

\begin{figure*}[t]
\centering
\includegraphics[width=0.88\textwidth] {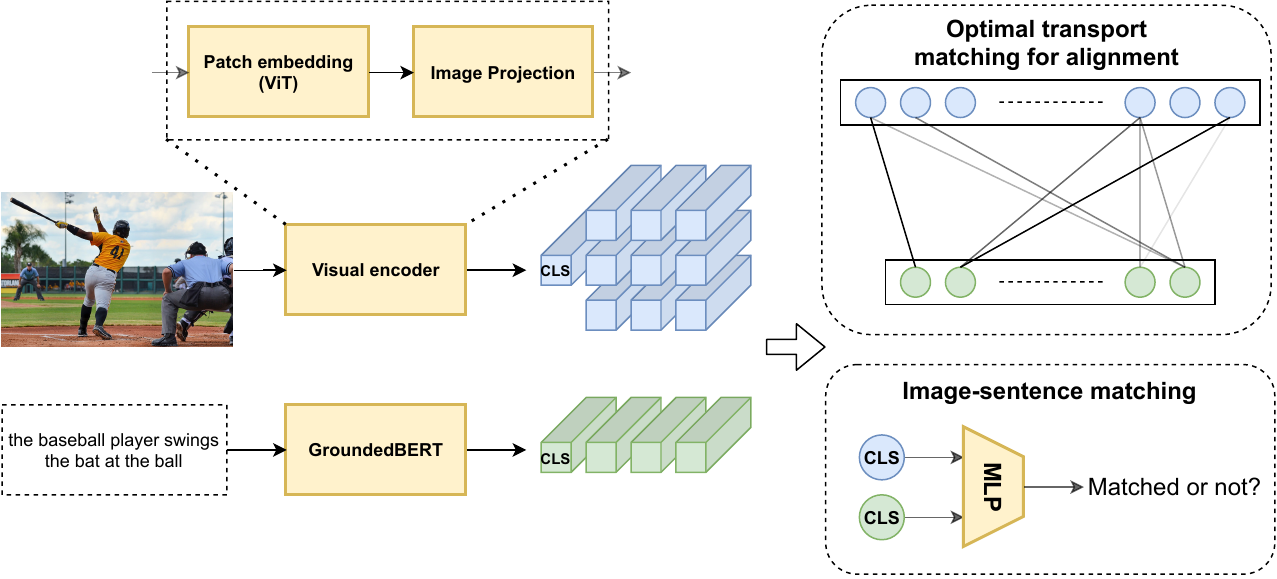}
\caption{ Implementation of our training framework. The framework consists of two parrallel pipeline for visual and text, finally, the whole model is train with two objectives: Image-sentence matching and Optimal transport matching for alignment.}
\vspace{-5pt}
\label{fig:loss_OT}
\end{figure*}

In this section, we introduce the details of our proposed GroundedBERT. As shown in Figure \ref{fig:vslm}, our model consists of two components: a language encoder and a visual grounding module. The complete framework is illustrated in Figure \ref{fig:loss_OT} where two objectives are introduced.

\subsection{Language encoder}

We use BERT \cite{devlin2019bert} as the language encoder. Given an input sentence $s = ( w_1, \ldots, w_n )$, we use the pretrained BERT model to contextually embed the discrete tokens ${w}_i$'s into hidden-output vector $\bm{h}_i$'s:

\vspace{-7pt}
\begin{align}
    \bm{h}_1, \bm{h}_2, \ldots, \bm{h}_l = \mathit{BERT}(w_1, w_2, \ldots, w_l)
\end{align} 
\vspace{-7pt}

\noindent where $\bm{h}_i = ( {h}^{1}_i, {h}^{2}_i, \ldots, {h}^{L}_i )$, ${h}^{l}_i$ is the hidden state of token $i$ at layer $l$ of the Transformer.

\subsection{Visual grounding}
\label{sec:gbert}

\paragraph{Visual grounding module}

The visual grounding module is a multi-layer perceptron to transform the contextual representation of each token in the sentence into the (visual) ground embedding. 

We take the hidden states of $k$ final Transformer layers ${h}^{L-k+1}_i, $ ${h}^{L-k+2}_i, \ldots, {h}^{L}_i$ and concatenate them as the input for the visual grounding module. 

\vspace{-7pt}
\begin{align}
    &\tilde{h}_i =  [{h}^{L-k+1}_i, {h}^{L-k+2}_i, \ldots, {h}^{L}_i] \\
    &g_i =\mathit{MLP}_{VG}(\tilde{h}_i)
\end{align}
\vspace{-7pt}

\noindent where $g_i$ is (visual) ground embedding of token $i$, $[{h}^{L-k+1}_i, {h}^{L-k+2}_i,$ $ \ldots, {h}^{L}_i]$ is the concatenation of hidden states of token $i$ from layer $L-k+1$ to $L$, VG stands for Visual Grounding.

\paragraph{Visual-Textual Embedding}

The textual embedding is the final hidden state of the language encoder. The ground embedding are concatenated to this textual embedding to form a unified visual-textual embedding of the token in the sentence.

\vspace{-7pt}
\begin{align}
    t_i &= [{h}^{L}_i, g_i] 
\end{align}
\vspace{-7pt}

\noindent where $t_i$ is the visual-textual embedding vector of the $i$-th token, which we take as the final representation of the token using our GroundedBERT model, $[{h}^{L}_i, g_i]$ is the concatenation of the final hidden state ${h}^{L}_i$ and the ground embedding $g_i$.

\subsection{Visual encoder}
\label{sec:visualencoder}
\paragraph{Patch embedding}

Instead of using traditional convolution-based architectures for visual feature extraction \cite{kiela2018learning, bordes2020incorporating, vokenization}, we use Vision Transformer (ViT)~\cite{ViT}. Let $img$ be the input image having size of $(c, w, h)$ which stands for the number of channels, width, and height of the image. Image $img$ goes through the ViT to get a global feature vector $\tilde{v}_{CLS}$ and $m$ patch embeddings $\tilde{v}_{1}, \dots,  \tilde{v}_{m}$.

\vspace{-7pt}
\begin{align}
    &\tilde{v}_{CLS}, \tilde{v}_{1}, \dots, \tilde{v}_{m} = \mathit{ViT}(img)
\end{align}
\vspace{-7pt}

\paragraph{Image projection}

We use a multi-layer perceptron to project the feature vector $\tilde{v}_i$ of each patch to the grounded space and represent visual context learned from visual features.

\vspace{-7pt}
\begin{align} 
    &v_{CLS}, v_{1}, ..., v_{m} = \mathit{MLP}_{prj}(\tilde{v}_{CLS}, \tilde{v}_{1}, ... \tilde{v}_{m} )
\end{align} 
\vspace{-7pt}

\noindent where $v_{CLS}$ is the global embedding of the input image, $v_{i}$'s are the patch embeddings and \textit{prj} stands for (image) projection.

\subsection{Training}
\label{sec:train}

In this section, we introduce two different optimization objectives: Image-sentence matching for global matching and Optimal transport matching for alignment between local features.

\paragraph{Image-sentence Matching}

The Image-sentence Matching task is inherited from the Image-text Matching task from many vision-and-language pretraining literatures mentioned in Section \ref{sec:related}. Learning how to perform well on this task will encourage the model to better find the relationship between the textual information and the visual signal in a global sense.

From each modality, we take a vector as its global representation. For the vision side, we use the global feature vector $v_{CLS}$ from ViT. For the language side, we use the visual-textual embedding of the CLS token. We concat these two vectors before feeding into a fully connected layer with sigmoid activation to make the binary prediction of whether the sentence describes the image.

\vspace{-7pt}
\begin{align}
    \hat{y} = \sigma(\mathit{FC}([v_{CLS}, t_{CLS}]))
\end{align}
\vspace{-7pt}

\begin{algorithm}[!t]
\caption{Computing Optimal Transport.}
\label{alg:ot}
\begin{algorithmic}[1]
\STATE {\bfseries Input:} \footnotesize{$\Cmat \in \mathbb{R}^{m \times n}$, $\av \in \mathbb{R}^m$, $\bv \in \mathbb{R}^n$, $\beta$, $iter$}
\STATE $\boldsymbol{\sigma}=\bm{1}_n / n$, $\Tmat = \bm{1}_m \bm{1}_n^\top$
\STATE $\Amat = \exp \left( -\frac{\Cmat}{\beta} \right)$
\FOR{$t = 1, 2, 3, \ldots, iter$}
    \STATE \footnotesize{// all division operations are element-wise}
    \STATE $\Qmat = \Amat \odot \Tmat$ \footnotesize{// $\odot$ is the Hadamard product}
    \STATE $\boldsymbol{\delta} = \frac{\av}{\Qmat{\boldsymbol{\sigma}}}$, $\boldsymbol{\sigma} = \frac{\bv}{\Qmat^\top\boldsymbol{\delta}}$
    \STATE $\Tmat = \text{diag}(\boldsymbol{\delta})\Qmat\text{diag}(\boldsymbol{\sigma})$
\ENDFOR
\STATE $\mathcal{D}=\langle \Cmat, \Tmat\rangle$ // \footnotesize{$\langle \cdot, \cdot \rangle$ is the Frobenius dot-product}  
\STATE Return $\Tmat$, $\mathcal{D}$ 

\end{algorithmic}
\end{algorithm}

\begin{algorithm}[!t]
\caption{Computing Partial Optimal Transport.}
\label{alg:pot}
\begin{algorithmic}[1]

\STATE {\bfseries Input:} \footnotesize{$\Cmat \in \mathbb{R}^{m \times n}$, $\av \in \mathbb{R}^m$, $\bv \in \mathbb{R}^n$, $\beta$, $s$, $iter$}

\STATE $\Tmat = \exp \left( -\frac{\Cmat}{\beta} \right)$
\STATE $\Tmat = \frac{s}{\bm{1}_n^\top \Tmat \bm{1}_m} \Tmat$
\FOR{$t=1, 2, 3, \ldots, iter$}
    \STATE \footnotesize{// all division operations are element-wise}
    \STATE $\knumv{a} = \min \left( \frac{\av}{\Tmat\bm{1}_n}, \bm{1}_m \right)$
    \STATE $\Tmat_{a} = \text{diag}(\knumv{a}) \Tmat $
    \STATE $\knumv{b} = \min \left( \frac{\bv}{\Tmat_{a}^\top \bm{1}_m}, \bm{1}_n \right)$
    \STATE $\Tmat_{b} = \text{diag}(\knumv{b}) \Tmat_{a}$
    \STATE $\Tmat = \frac{s}{\bm{1}_n^\top \Tmat_{b} \bm{1}_m}  \Tmat_{b}$ 
\ENDFOR
\STATE $\mathcal{D}=\langle \Cmat, \Tmat\rangle$  // \footnotesize{$\langle \cdot, \cdot \rangle$ is the Frobenius dot-product} 
\STATE Return $\Tmat$, $\mathcal{D}$ 
\end{algorithmic}
\end{algorithm}

\begin{table*}[ht]
\caption{
Downstream task results of BERT, Vokenization \cite{vokenization} and our GroundedBERT, we conduct the experiments on BERT-base architecture. MRPC results are F1 score, STS-B results are Pearson correlation, SQuAD v1.1 and v2.0 results are exact matching and F1 score. The results, which outperform the other one are marked in bold, are all scale to range 0-100. The $\Delta_{base}$ and $\Delta_{Vok}$ columns show the difference between our model and the baseline, and the Vokenization method respectively.
}
\centering
\resizebox{0.85\textwidth}{!}{
\begin{tabular}{@{}|l|c|cc|cc|@{}}
\toprule
\textbf{Task} & Our & BERT-base & $\Delta_{base}\uparrow$ & Vokenization\footnote{The results are from the Vokenization \cite{vokenization} paper} & $\Delta_{Vok}\uparrow$ \\ 
\midrule
CoLA & \textbf{60.95} & 54.68  & 2.41 & \_ & \_\\
MNLI & \textbf{84.15} & 83.48 & 0.84 & 82.6 & 1.55\\
MNLI-MM & \textbf{84.54} & 84.05 & 0.83 & \_ & \_ \\
MRPC & \textbf{89.25} & 88.82 & 0.74 & \_ & \_\\
QNLI & \textbf{91.43} & 91.37 & 0.5 & 88.6 & 2.83\\
RTE & \textbf{72.56} & 67.87 & 3.6 & \_ & \_\\
SST-2 & \textbf{93.12} & 92.43 & 0.57 & 92.2 & 0.92\\
STS-B & \textbf{89.88} & 89.00 & 0.84 & \_ & \_\\
\midrule
SQuADv1.1 & 78.49/86.62 & 78.10/86.31& 0.39/0.32 & \textbf{78.8/86.7} & -0.31/-0.08 \\
SQuADv2.0 & \textbf{70.69/73.92} & 67.92/71.08 & 2.77/2.84 & 68.1/71.2 & 2.59/2.72 \\
\bottomrule
\end{tabular}
\label{table:results}
\vspace{-5pt}
}
\end{table*}

\noindent where $\hat{y}$ is predicted probability, $\sigma(x) = [1 + \exp(-x)]^{-1}$ is the sigmoid function, $[.,.]$ is the concatenation operation.




The negative pair is created by replacing the image with another randomly selected image from the training set. We apply the binary cross-entropy loss for optimization.

\vspace{-7pt}
\begin{equation}
    \mathcal{L}_{cls} = - y \log \hat{y} - (1 - y) \log (1-\hat{y})
\end{equation}
\vspace{-7pt}

\noindent where $y$ is the binary indicator, $y = 1$ if the image matches the sentence and $0$ otherwise.

\paragraph{Optimal transport for vision-language alignment}

To solve the alignment between language and vision, we use Optimal Transport (OT), specifically the Partial Optimal Transport (POT) variant. 

For each image, we have $m$ patch embeddings $\bm{v} = ( v_1, \dots, v_m )$. For each sentence, we have $n$ hidden representations of the words $\bm{t} = ( t_1, \dots, t_n \}$. We consider these two collections as the supports of two empirical distributions with uniform weights. We then use OT to estimate the distance between these two distributions. 

Specifically, we compute the cost matrix $\Cmat$ where $c_{ij} = 1 - \cos \angle (v_i, t_j)$, or the cosine distance between the corresponding patch and word embedding. We also let $\av = \bm{1}_m / m$ and $\bv = \bm{1}_n / n$ be the two uniform weight vectors.

\begin{table}[t]
\caption{Task descriptions and statistics.}
\centering
\resizebox{0.46\textwidth}{!}{
\begin{tabular}{@{}l|ccc@{}}
\toprule
\textbf{Corpus} & \textbf{Train} & \textbf{Test} & \textbf{Metrics}\\
\midrule
\multicolumn{4}{c}{GLUE} \\
\midrule
CoLA & 8.5k & 1k & Matthews corr \\
MNLI & 393k & 20k & matched acc./mismatched acc. \\
MRPC & 3.7k & 1.7k & acc./F1 \\
QNLI & 105k & 5.4k & acc. \\
RTE & 2.5k & 3k & acc. \\
SST-2 & 67k & 1.8k & acc. \\
STS-B & 7k & 1.4k & Pearson corr.\\
\midrule
\multicolumn{4}{c}{SQUAD} \\
\midrule
SQUAD V1.1 & 87K & 10K & exact match/F1\\
SQUAD V2.0 & 130K & 11K & exact match/F1\\
\bottomrule
\end{tabular} }

\label{tab:tasks}
\vspace{-5pt}
\end{table}

The distance between the two modalities can be defined using the OT-based distance as

\vspace{-7pt}
\begin{equation}
    \begin{aligned}
    \mathcal{D}(\bm{v}, \bm{t}) = \min_{\Tmat}  \left \langle \Tmat, \Cmat \right \rangle_F \\
    \textrm{s.t.} \quad \Tmat\mathbf{1}_n = \av, \Tmat^\top\mathbf{1}_m = \bv, \Tmat \succeq \bm{0}_{m \times n} \\
    \end{aligned}
\end{equation}
\vspace{-7pt}

This formulation places constraints that all the mass from one distribution must be transported to the other distribution. We find, however, that this constraint is restrictive for the problem at hand, where the sentence describes only partially the corresponding image. Therefore, it is intuitively more apt to use the POT variant, described as follows. 

\vspace{-7pt}
\begin{equation}
    \begin{aligned}
    \mathcal{D}(\bm{v}, \bm{t}) = \min_{\Tmat} \left \langle \Tmat, \Cmat \right \rangle_F \\
    \textrm{s.t.} \quad \Tmat\mathbf{1} \preceq \av, \Tmat^\top\mathbf{1}_n \preceq \bv, \Tmat \succeq \bm{0}_{m \times n} \\
      \mathbf{1}_m^\top \Tmat \bm{1} = s
    \end{aligned}
\end{equation}
\vspace{-7pt}

\noindent where $s$ is the total amount of mass to be transported. In our implementation, $s$ is set as the total uniform weight vector of text.

We use sinkhorn-based algorithms to calculate the transportation plan $\Tmat$ and the OT-based distance. Algorithm \ref{alg:ot} and Algorithm \ref{alg:pot} are for OT and POT, respectively. The average distance $\mathcal{D}$ for every matching pair of sentence and image will be minimized, while non-matching pair distance will be maximized. Formally, the alignment loss will be:

\vspace{-7pt}
\begin{equation}
    \mathcal{L}_{align} = \sum_{\bm{t}, \bm{v^+}, \bm{v^-} \in S} \left[ \mathcal{D}(\bm{v^+}, \bm{t}) - \mathcal{D}(\bm{v^-}, \bm{t}) \right]
\end{equation}
\vspace{-7pt}

\noindent where $S$ is the given dataset, $\bm{v^+}$ and $\bm{v^-}$ are the matching and non-matching image respectively corresponding to the sentence $t$. The procedure to pick the negative image is similar to in the Image-Sentence Matching task.

{\footnotesize

\begin{table*}[t]
\caption{
Downstream task results of different vision and language pretrained model.
}
\centering
\resizebox{0.84\textwidth}{!}{
\begin{tabular}{@{}|l|cccccc|@{}}
\toprule
\textbf{Task} & LXMERT & VisualBERT & VL-BERT & ViLBERT & Oscar & GroundedBERT \\ 
\midrule
CoLA & 15.76 & 45.14 & 57.01 & 56.05 & 41.21 & \textbf{57.09} \\
MNLI & 35.44 & 80.68 & 81.18 & 81.29 & 76.64 & \textbf{84.32}  \\
MNLI-MM & 35.22 & 80.96 & 81.38 & 81.02 & 76.67 & \textbf{84.88} \\
MRPC & 80.64 & 87.36 & 87.76 & 86.95 & 80.58 & \textbf{89.56}\\
QNLI & 50.54 & 87.39 & 89.20 & 86.95 & 50.54 & \textbf{91.87} \\
RTE & 52.71 & 66.43 & 62.09 & 70.40 & 55.96 & \textbf{71.47} \\
SST-2 & 82.11 & 88.88 & 88.88 & 90.14 & 87.61 & \textbf{93.00}\\
STS-B & 42.23 & 90.03 & 89.48 & 89.98 & 71.45 & \textbf{89.84}\\
\midrule
SQuADv1.1 & 9.39/17.65 & 68.51/77.71 & 72.62/81.30 & 72.95/81.35 & 21.77/32.20 & \textbf{78.49/86.62}\\
SQuADv2.0 & 46.52/47.04 & 59.17/62.53 & 62.38/65.63 & 63.36/66.56 & 45.31/46.77 & \textbf{70.69/73.92}\\
\bottomrule
\end{tabular} }

\label{table:vlmresults}
\vspace{-5pt}
\end{table*}
}

\section{Experimental Setup}
\label{sec:setup}

\subsection{Datasets}

\paragraph{Training}

We use MS COCO~\cite{mscoco} and Visual Genome~\cite{krishnavisualgenome} image captioning datasets as the training data for image projection and Visual grounding module.

\paragraph{Evaluation}





\begin{table*}[t]
\caption{
Downstream task results and comparison of our GroundedBERT without training the Visual grounding module. The first two rows report the fine-tuned results of our model without training with the visual grounded datasets, while the last 4 rows show the results of our approaches on both OT and POT.
}
\centering
\resizebox{0.9\textwidth}{!}{%
\begin{tabular}{@{}|l|cccccccccc|@{}}
\toprule
Dimension                 & \small CoLA& \small MNLI & \small MNLI-MM & \small MRPC & \small QNLI & \small RTE & \small SST-2 & \small STS-B & \small SQuAD V1.1 & \small SQuAD V2.0 \\ \midrule
$\text{64}_{wo}$ & 54.37 & 83.27 & 84.47 & 88.11 & 90.78 & 69.18 & 91.71 & 88.97 & 77.87/86.2 & 67.78/70.97 \\
$\text{128}_{wo}$ & 53.59 & 83.78 & 84.04 & 88.63 & 91.45 & 69.53 & 91.12 & 89.3 & 77.98/86.03  & 68.18/71.45 \\
\midrule
$\text{64}_{OT}$ & 58.30 & 84.35 & 84.64 & 88.71 & 91.58 & 70.40 & 92.43 & 89.32 & 78.14/86.42 & 69.00/72.24 \\
$\text{128}_{OT}$ & 59.1 & 84.49 & 84.76 & 88.81 & 91.61 & 69.31 & 92.78 & 89.75 & 78.21/86.46 & 68.6/71.94\\
\midrule
$\text{64}_{POT}$ & 60.95 & 84.15 & 84.54 & 89.25 & 91.43 & 72.56 & 93.12 & 89.88 & 78.49/86.62 & 70.69/73.92\\
$\text{128}_{POT}$ & 57.77 & 84.06 & 84.3 & 88.93 & 91.31 & 70.04 & 92.32 & 89.86 & 78.27/86.45 & 69.53/72.88\\
\bottomrule

\end{tabular}}

\label{table:ablation2}
\vspace{-5pt}
\end{table*}


After training process, we finetune and evaluate our model on GLUE~\cite{wang2018glue}, SQuAD 1.1 ~\cite{rajpurkar2016squad}, and SQuAD 2.0 datasets \cite{rajpurkar2018know}. In GLUE dataset, we evaluate our model on various tasks over 7 corpora: CoLA~\cite{cola}, MNLI~\cite{mnli}, MRPC~\cite{mrpc}, QNLI~\cite{rajpurkar2016squad}, RTE, SST-2~\cite{sst}, STS-B~\cite{sts}. The statistics of  datasets are given in table \ref{tab:tasks}.

\subsection{Evaluation tasks and metrics}

All tasks are single sentence or sentence pair classification except STS-B, which is a regression task. MNLI has three classes, all other classification tasks are binary classification. The evaluation tasks are also various: question answering (QNLI, SQUAD), acceptability (CoLA), sentiment (SST-2), paraphrase (MRPC), inference (MNLI, RTE, QNLI). The metric of each task is shown in table \ref{tab:tasks}. For MRPC, we report F1 score. For STS-B, we report Pearson correlation. For both SQuAD, we report exact matching and F1 score.


\subsection{Implementation}
\label{sec:imple}

We use BERT-base-uncased as the language model and vit base patch16 224 for the visual encoder. We load the BERT weight pretrained on Bookcorpus and Wikipedia from Pytorch framework Huggingface, and load the ResNeXt weight pretrained on ImageNet. The Language encoder and Patch embedding extraction are frozen, we just train the Image projection and Visual grounding module based on the contextual representation and image feature map. Both modules are multi-layer perceptron with 1 hidden layers and apply relu activation. We set the MLP final output dimension in set $64,128$ for evaluating how visual information impact on the textual-visual representation in Sec \ref{sec:vdim}. Our model is trained with a learning rate $l_r = 1e^{-4}$ in $12$ epochs using AdamW~\cite{loshchilov2018fixing} as optimizer, we set batch size of 512 on 1 V100 GPU and train for 3-4 days.

\begin{table*}[t]
\caption{
Downstream task results of BERT and our GroundedBERT with different learning rates on GLUE.
}

\small
\centering
\resizebox{0.86\textwidth}{!}{%
\begin{tabular}{@{}|l|c|cccccccc|@{}}
\toprule
Model & LR & \small CoLA & \small MNLI & \small MNLI-MM & \small MRPC & \small QNLI & \small RTE & \small SST-2 & \small STS-B \\ 
\midrule
\multirow{4}{*}{BERT-base} & $\text{2e-5}$ & 53.13 & 83.48 & 83.68 & 86.78 & 91.37  & 64.62 & 92.43 & 88.74 \\
& $\text{3e-5}$ & 54.68 & 82.98 & 84.05 & 87.25 & 90.83 & 66.79 & 92.43 & 88.44 \\
& $\text{4e-5}$ & 53.80 & 83.25 & 83.76 & 88.82 & 90.76 & 67.87 & 92.09 & 89.00 \\
& $\text{5e-5}$ & 52.85 & 82.52 & 82.72 & 88.72 & 90.06 & 67.15 & 92.09 & 88.60 \\
\midrule

\multirow{4}{*}{Our OT + Classifier} & $\text{2e-5}$ & 55.73 & 84.35 & 84.64 & 87.15 & 90.54 & 68.23 & 91.74 & 89.32 \\
& $\text{3e-5}$ & 58.30 & 83.88 & 84.13 & 87.46 & 91.58 & 66.79 & 92.09 & 88.63 \\
& $\text{4e-5}$ & 55.22 & 83.24 & 83.79 & 88.71 & 90.33 & 65.70 & 92.43 & 88.59 \\
& $\text{5e-5}$ & 54.48 & 81.63 & 82.32 & 88.56 & 90.33 & 70.40 & 90.37 & 88.97\\
\midrule

\multirow{4}{*}{Our POT + Classifier} & $\text{2e-5}$ & 58.44 & 84.15 & 84.52 & 88.13 & 91.43 & 66.79 & 93.12 & 89.88 \\
& $\text{3e-5}$ & 60.95 & 83.43 & 84.54 & 89.25 & 90.96 & 70.04 & 92.20 & 88.75 \\
& $\text{4e-5}$ & 58.07 & 82.36 & 83.14 & 87.69 & 90.76 & 72.56 & 92.09 & 89.11 \\
& $\text{5e-5}$ & 52.50 & 82.59 & 83.08 & 88.51 & 90.32 & 68.59 & 91.51 & 88.71 \\

\bottomrule

\end{tabular}}

\label{table:ablationLR}
\vspace{-5pt}
\end{table*}

\begin{table*}[t]
\caption{
Downstream task results different approaches when training the Visual grounding module.}
\centering
\resizebox{0.95\textwidth}{!}{%
\begin{tabular}{@{}|l|l|cccccccccc|@{}}
\toprule
Optimal transport? & Classification & \small CoLA& \small MNLI & \small MNLI-MM & \small MRPC & \small QNLI & \small RTE & \small SST-2 & \small STS-B & \small SQuAD V1.1 & \small SQuAD V2.0\\ \midrule
No & Yes & 58.05 & 83.93 & 84.12 & 88.64 & 91.20 & 69.31 & 92.78 & 89.37 & 78.34/86.42 & 70.14/73.71\\
Classical & No & 60.85 & 84.17 & 84.85 & 88.68 & 90.87 & 72.20 & 92.43 & 89.45 & 77.46/85.88 & 67.67/71.4\\
Partial & No & 58.20 & \textbf{84.38} & \textbf{84.88} & \textbf{89.25} & 90.99 & 69.31 & 92.66 & 89.27 & 77.99/86.35 & 68.29/71.65\\
Classical & Yes & 58.30 & 84.35 & 84.64 & 88.71 & \textbf{91.58} & 70.40 & 92.43 & 89.32 & 78.14/86.42 & 69.00/72.24 \\
Partial & Yes & \textbf{60.95} & 84.15 & 84.54 & \textbf{89.25} & 91.43 & \textbf{72.56} & \textbf{93.12} & \textbf{89.88} & \textbf{78.49/86.62} & \textbf{70.69/73.92}\\

\bottomrule
\end{tabular}}


\label{table:abinformation}
\vspace{-5pt}
\end{table*}

\section{Experimental Results}
\label{sec:result}

\subsection{Compared to the baseline models}
The fine-tune results on 9 different natural-language tasks are reported in Table \ref{table:results}. We compare our GroundedBERT with the BERT-base as the language encoder to the BERT-base and Vokenization baseline respectively. Our GroundedBERT outperforms the baselines on all down-stream tasks. Specifically, we achieve an improvement from 0.5 to 3.6 score on BERT-base. Compared to Vokenization, we also achieve higher on most tasks, except SQuADv1.1. This shows that our grounded language model representation can capture more useful information for language understanding without changing the original language model.

\subsection{Compared to other vision-and-language pretrained models}

To prove the effectiveness of our proposed grounded language learning approach, we compare it with the following state-of-the-art vision-and-language pretrained models.

\begin{itemize}[leftmargin=*]
    \item \textbf{LXMERT} \cite{tan2019lxmert} consists of two single-modal and one cross-modal Transformer to connect vision and language semantics.
 
    \item \textbf{VisualBERT} \cite{li2019visualbert} consists of a stack of Transformer layers that implicitly align elements of an input text and regions in an associated input image with self-attention.
   
    \item \textbf{VL-BERT} \cite{su2019vl} uses Transformer model as the backbone, and extends it to input both visual and linguistic embedded features.
    
    \item \textbf{ViLBERT} \cite{lu2019vilbert} extends the BERT architecture to a multi-modal two-stream model and process both visual and textual inputs.
    
    \item \textbf{Oscar} \cite{li2020oscar} uses object tags detected in images as additional points to ease the learning of alignments between text and image.
\end{itemize}

We also fine-tune all models on 9 different natural-language tasks of GLUE and SQuAD datasets. To have a fair comparison, all models are initialized with the pretrained BERT weights, except LXMERT that is pretrained from scratch. As shown in Table \ref{table:vlmresults}, the finetuning results on our model consistently outperform other pretrained models in all tasks. The results show that finetuning the BERT model will make it forget the original knowledge learned from a huge language corpus.



\section{Analysis}

\subsection{The impact of visual grounding}
\label{sec:vdim}
To understand the impact of visual grounding on text representation, we train GroundedBERT without using visual information and the weights of the Visual grounding module are randomly initialized. The results in Table \ref{table:ablation2} show that the visual information has the significant contribution in the language grounding and is beneficial to the textual representation. We also study the impact of the contribution visual grounding with different visual embedding dimensions. Since the dimension of the hidden representation of language encoder, which is BERT, is fixed depend on its configuration, we can setup the dimension of additional visual information flexibly.


\subsection{Different learning rates on GLUE}

Following the setting in BERT \cite{devlin2019bert} on GLUE tasks, we also conduct additional experiments with more runs on different learning rates similar to the BERT paper. The learning rates are also similarly set to be $\{2,3,4,5\}e-5$ to have a fair comparison. The results in Table \ref{table:ablationLR} show that our model consistently outperforms the baseline in all datasets for all learning rates.

\subsection{Different training strategy}

We conduct the experiments on GroundedBERT trained with different settings: Optimal transport and Classifier. Table \ref{table:abinformation} reports evaluations of our model on GLUE and SQUAD on 5 different approaches, i.e., only Classifier, only Classical OT (OT), only Partial OT, Classifier + OT and Classifier + POT. The results show that the combination of Classifier and Partial OT achieves the highest score in most tasks, while Partial OT perform better than Classical OT in both combination with Classifier or not.

\section{Conclusion}
\label{sec:conclusion}

In this paper, we propose GroundedBERT as a grounded language learning model that incorporates visual information into BERT representation. We introduce the visual grounding module to capture the visual information which is later joined with the text representation to create a unified visual-textual representation. Our model significantly outperforms the baseline language models on various language tasks of the GLUE and SQuAD datasets.


\begin{acks}
This research is supported by the National Research Foundation, Singapore under AI
Singapore Programme, AISG Award No: AISG2-
TC-2022-005
\end{acks}

\bibliographystyle{acmmm23}
\bibliography{acmmm23}

\end{document}